\documentclass[lettersize,journal,twoside]{IEEEtran}

\usepackage{amsmath,amsfonts,amssymb}
\usepackage{mathtools}
\usepackage{algorithmic}
\usepackage{algorithm}
\usepackage{array}
\usepackage[caption=false,font=normalsize,labelfont=sf,textfont=sf]{subfig}
\usepackage{textcomp}
\usepackage{stfloats}
\usepackage{url}
\usepackage{graphicx}
\usepackage{cite}
\usepackage{booktabs}
\usepackage{hyperref}

\let\oldthanks\thanks
\renewcommand{\thanks}[1]{\oldthanks{\vspace{-0.2pt}#1}}

\newcommand{\sM}{{\mathcal{M}}}
\newcommand{\sS}{{\mathcal{S}}}
\newcommand{\sA}{{\mathcal{A}}}
\newcommand{\sH}{{\mathcal{H}}}

\begin{document}

\title{Abstract Sim2Real through Approximate Information States}

\author{Yunfu~Deng,~Yuhao~Li,~and~Josiah~P.~Hanna
\thanks{Manuscript received: September 16, 2025; Revised: January 29, 2026; Accepted: March 25, 2026.}%
\thanks{This paper was recommended for publication by Editor Aniket Bera upon evaluation of the Associate Editor and Reviewers' comments.This work took place in the Prediction and Action Lab (PAL) at the University of Wisconsin -- Madison.
PAL research is supported by NSF (IIS-2410981) and the Wisconsin Alumni Research Foundation.}%
\thanks{Yunfu Deng and Josiah P. Hanna are with the Department of Computer Sciences, University of Wisconsin--Madison, Madison, WI 53706 USA (e-mail: yunfu.deng@wisc.edu, jphanna@cs.wisc.edu).}%
\thanks{Yuhao Li performed this work while at the University of Wisconsin--Madison. He is now with the Manning College of Information and Computer Sciences, University of Massachusetts Amherst, Amherst, MA 01003 USA (e-mail: yuhaoli@umass.edu).}%
\thanks{Digital Object Identifier (DOI): see top of this page.}}

\renewcommand{\markboth}[2]{\def\leftmark{#1}\def\rightmark{#2}}
\markboth{IEEE ROBOTICS AND AUTOMATION LETTERS. PREPRINT VERSION. ACCEPTED MARCH, 2026}%
{Deng \textit{et al.}: Abstract Sim2Real through Approximate Information States}

\maketitle

\begin{abstract}
In recent years, reinforcement learning (RL) has shown remarkable success in robotics when a fast and accurate simulator is available for a given task. When using RL and simulation, more simulator realism is generally beneficial but becomes harder to obtain as robots are deployed in increasingly complex and widescale domains. In such settings, simulators will likely fail to model all relevant details of a given target task and this observation motivates the study of sim2real with simulators that leave out key task details. In this paper, we formalize and study the abstract sim2real problem: given an abstract simulator that models a target task at a coarse level of abstraction, how can we train a policy with RL in the abstract simulator and successfully transfer it to the real-world? Our first contribution is to formalize this problem using the language of state abstraction from the RL literature. This framing shows that an abstract simulator can be grounded to match the target task if the grounded abstract dynamics take the history of states into account. Based on the formalism, we then introduce a method that uses real-world task data to correct the dynamics of the abstract simulator. We then show that this method enables successful policy transfer both in sim2sim and sim2real evaluation.
\end{abstract}


\section{Introduction}
\IEEEPARstart{R}{einforcement} learning (RL) has demonstrated remarkable success across diverse application domains, from game playing \cite{wurman2022outracing} to robotic manipulation \cite{andrychowicz2020learning}, navigation \cite{wijmans2019dd}, and locomotion \cite{hwangbo2019learning}. Despite these achievements, deploying RL in complex, real-world scenarios remains non-trivial due to a combination of expensive data collection, partial observability, and intricate physical dynamics.
 
Simulators offer a safer and less costly alternative to real-world learning, but standard sim2real methods---including domain randomization \cite{peng2018sim}, system identification, and learned dynamics corrections---assume that the simulator and the target domain share the same state-action space and differ only in dynamics parameters \cite{yoon2023comparative}. These methods address parametric mismatch but may not apply when the simulator operates over a different and more abstract state representation than the target robot---a common scenario when constructing a high-fidelity simulator is impractical. Noorani et al. \cite{noorani2025abstraction} argue that reliance on high-fidelity simulation leads to overfitting to simulator-specific dynamics and that abstract, lower-fidelity simulators may instead enable more generalizable autonomy.
 
Motivated by these observations, researchers have turned to understanding how to use much more simplified and abstract simulators for sim2real \cite{truong2023rethinking,hofer_perspectives_2020,Labiosa_Multi-Robot_2025,Labiosa_Reinforcement_2025}. More abstract simulators can significantly speed up experimentation and simplify the modeling process, making RL more accessible and efficient to develop \cite{Labiosa_Multi-Robot_2025,truong2023rethinking}.
However, a highly simplified simulator runs the risk of ignoring essential task-relevant dynamics in the real world, leading to policies that fail when transferred.
Furthermore, the abstract sim2real problem has yet to be mathematically formalized in the literature, and this gap hinders the development of theoretically grounded abstract sim2real methods.

With this motivation in mind, in this paper, we aim to answer the question: 
\begin{center}
    \textit{``Can we use real-world data to modify an abstract simulator for effective sim2real transfer of RL-trained behaviors for a real robot?"}
\end{center}
Toward answering this question, we provide the first (to the best of our knowledge) formalization of the abstract sim2real problem using the notion of a state abstraction from the RL literature.
Using this formalism, we identify that the key to abstract simulator grounding for successful policy transfer is to learn simulator corrections and policies as functions of abstract state and action histories to mitigate the effect of partial observability induced by abstraction.
History-based grounding enables the grounded abstract simulator to implicitly account for unmodelled real-world dynamics.
We leverage this insight to develop a new method, ASTRA (Augmented Simulation with self-predicTive abstRAction), that uses a small amount of real-world data to ground an abstract simulator.
We validate this method through sim2real tasks with the humanoid NAO robot and sim2sim experiments in navigation and humanoid locomotion, where we show that it enables successful abstract sim2real transfer where other baselines (including a strong domain randomization approach) \cite{huang2023went} fail.

\section{Related Work}

Sim2real transfer has been extensively studied in robotics; see Zhao et al.\ \cite{zhao_sim--real_2020} for a more complete survey. 
%
In this section, we focus on the most closely related literature on abstraction in sim2real, simulator grounding methods, and state abstraction in RL.

\subsection{Abstraction and Sim2Real}
The sim2real research community has acknowledged the importance of developing the use of capability of robots that can learn with abstract simulators of the world \cite{hofer_perspectives_2020,noorani2025abstraction}.
In addition to the practical motivation that an abstract simulator is easier to specify, evidence from psychology suggests that humans seamlessly plan actions with abstract models \cite{ho_people_2022}, raising the question of how robots might also base their actions on abstract models.
Several works have demonstrated the possibility and even advantages of abstract simulators compared to high-fidelity simulators.
Nachum et al.\ showed that a high-level policy for robot navigation can be trained in a high-fidelity simulator but only using an abstract state representation as input \cite{nachum2019multi}.
M{\"u}ller et al. show that abstraction can aid sim2real transfer in autonomous driving \cite{muller2018driving}.
Jain et al. improve learning of visual navigation policies by first training in an abstract grid simulator \cite{jain2021gridtopix}.
Truong et al. found that reducing a simulator's fidelity (switching from a dynamics to kinematic motion model) enables improved transfer of visual navigation policies, particularly when the wall-clock time of training is limited \cite{truong2023rethinking}.
Our work differs from these studies in that we provide the first formal description of the abstract sim2real problem and then develop a new abstract simulator grounding method based on this formalism.
%
Cutler et al.\ also formalize the notion of multi-fidelity simulators and develop a method for transferring knowledge between simulators \cite{cutler2014reinforcement}, but focus on value approximation rather than state space mismatches.
In contrast, our formal model shows that abstraction induces partial observability, necessitating history-based grounding methods.

\subsection{Simulator Grounding}
The methods we introduce are most closely related to existing methods for sim2real that ground a simulator's dynamics to more closely match real-world dynamics.
This class of methods includes system identification, by which the parameters of a simulator are tuned based on real-world experimental trials \cite{astrom_system_1971,armstrong_finding_1987}.
Many recent works in the sim2real literature have proposed using real-world data to learn corrections to a given simulator \cite{bousmalis2018using,golemo2018sim,hanna2021grounded,karnan_reinforced_2020}.
These works focus on how to apply corrections and how to learn them with limited data.
We focus on similar challenges but differ from prior work by focusing on these challenges in the context of highly abstract simulators where the state abstraction $\phi$ induces partial observability that standard grounding methods cannot address.

\subsection{Reinforcement Learning with Abstraction}

Our work uses the theory of state abstraction from the RL literature to formalize and identify methods for the abstract sim2real problem.
While there is substantial work in state-abstraction (see Abel \cite{abel_theory_2022} for a survey), abstract sim2real is most closely related to the use of abstraction for model-based RL \cite{jiang_abstraction_2015,chaudhari_abstract_2024}.
The key difference between these works and abstract sim2real is that sim2real starts with a given model (i.e., simulator) that is specified with domain knowledge.
State abstraction is known to induce partial observability when applied to states in an MDP \cite{allen_learning_2021}.
When states cannot be fully observed (i.e., in POMDPs), integrating historical information becomes critical \cite{littman2001predictive,hausknecht2015deep}.

\section{Abstract Sim2Real}
\label{sec:abstract-sim2real}
In this section, we formalize the abstract sim2real problem. After establishing notation for standard sim2real, we define the abstract sim2real setting where the simulator's state representation is reduced (i.e., abstracted) from the full real-world state.
We show that this difference fundamentally changes the transfer problem: abstraction induces partial observability, necessitating history-based methods for simulator grounding.

\subsection{Preliminaries}
\paragraph{Reinforcement Learning}
We formalize an RL task as a Markov decision process (MDP),
\(
\mathcal{M} = \bigl(\mathcal{S}, \mathcal{A}, P, r, \gamma\bigr),
\)
in which $\mathcal{S}$ denotes the set of states, $\mathcal{A}$ the set of actions, 
$P\colon \mathcal{S}\times\mathcal{A}\to\Delta(\mathcal{S})$ the transition dynamics, 
$r\colon \mathcal{S}\times\mathcal{A}\to\mathbb{R}$ the reward function, 
and $\gamma\in[0,1)$ the discount factor.
A policy \(\pi : \mathcal{S} \to \Delta(\mathcal{A})\) maps states to action distributions, 
with objective to maximize expected discounted return:
$
J_{\mathcal{M}}(\pi) \;=\; \mathbb{E}\Bigl[\sum_{i=0}^{\infty} \gamma^i \, r\bigl(s_i, a_i\bigr)\Bigr].
$

\paragraph{Sim2Real}
Let the target domain (real world) be represented by the MDP \(\mathcal{M}_t \coloneqq \bigl(\mathcal{S}, \mathcal{A}, P_t, r, \gamma\bigr)\), and the source domain (simulator) by \(\mathcal{M}_s = \bigl(\mathcal{S}, \mathcal{A}, P_s, r, \gamma\bigr)\).
The sim2real problem is to find a policy $\pi$ using RL in $\mathcal{M}_s$ that maximizes $J_{\mathcal{M}_t}(\pi)$.
The main challenge is that $P_s \neq P_t$ and consequently a policy that maximizes $J_{\mathcal{M}_s}$ may fail w.r.t.\ $J_{\mathcal{M}_t}$.
Most sim2real work explicitly or implicitly focuses on what we call \textit{parametric mismatch}: the source and target domain dynamics have approximately similar structure but differ in parameters. 
With parametric mismatch, the reality gap can be addressed through system identification \cite{tan_sim--real_2018}, simple, learned dynamics corrections \cite{hanna2021grounded}, or domain randomization \cite{peng2018sim}.

\subsection{Problem Formulation}

We define an abstract simulator as an MDP over an abstract state-space,
$
\mathcal{M}_s :=\left(\mathcal{S}^s, \mathcal{A}, P_s, r_s, \gamma\right),
$
where the state space \(\mathcal{S}^s\) is deliberately reduced from the real state space \(\mathcal{S}^t\). 
Two examples of abstract state spaces are either compressing states in $\sS^t$ to a finite set of discrete abstract states or to have $\sS^s$ be a subspace of $\sS^t$ (e.g., by dropping some dimensions of the real-world state).
We assume the mapping from real-world states to abstract simulator states is known and define it as: $\phi: \mathcal{S}^t \rightarrow \mathcal{S}^s$.\footnote{This assumption is mild in many robotic systems where high-level semantic state variables are extracted from existing perception and state-estimation pipelines.}
With slight abuse of notation, we also write $\phi: \sH^t \rightarrow \sH^s$ to denote applying $\phi$ to every state $s^t$ in a trajectory of target environment states, $h^t \in \sH^t$, so as to obtain a trajectory of abstract states, $h^s \in \sH^s$.
Note that the simulator's transition function, \(P_s\), is structurally different from \(P_t\) because $\phi$ merges or discards details that are present in \(\mathcal{S}^t\).
In this work, we will assume that the abstract simulator and real-world share the same action space, $\mathcal{A}$.
If action spaces differ (e.g., high‑level commands vs. low‑level torques), we assume access to an action transformation function, such as a learned low‑level policy, to map actions from \(\sM_s\) to \(\sM_t\).
As with standard sim2real, our objective is to use RL in $\sM_s$ to learn a policy, $\pi$, that maximizes $J_{\sM_t}(\pi).$

Having formalized abstract sim2real, we observe that it raises two primary challenges beyond the standard sim2real problem, both rooted in the fact that state abstractions generally induce partial observability \cite{allen_learning_2021}.
First, a policy $\pi: \sS^s \rightarrow \Delta(\sA)$ trained in the abstract simulator might be missing information critical for optimal control that is otherwise available in $\sS^t$.
For example, consider a 2D abstract simulator with state-space $\mathcal{S}^s = [x, y, v_x, v_y]$ that models a NAO bipedal robot with state $\mathcal{S}^t \in \mathbb{R}^{30}$ including joint configurations. The abstract state cannot distinguish between a biped with stable stance ready to navigate and one that has become unsteady due to a recent rapid change in direction. Both configurations map to identical values of $(x, y, v_x, v_y)$, yet they require fundamentally different actions. 

The second consequence of the partial observability induced by abstraction is that we cannot directly apply simulator grounding methods using experience from $\sM_t.$
To see this, consider if we have a trajectory, $s_0^t, a_0, s_1^t,...,s_T^t$, collected by running some policy in $\sM_t$.
Applying the abstraction, $\phi$, to each state in this trajectory results in a trajectory, $s_0^s, a_0, s_1^s,...,s_T^s$.
However, the Markov property, in general, fails to hold in terms of abstract states meaning that $\Pr(s_{i+1}^s|s_i^s, a_i) \neq \Pr(s_{i+1}^s|s_i^s, a_i, s_{i-1}^s, a_{i-1})$ in the target domain \cite{allen_learning_2021}.
This violation makes the standard grounding objective of trying to make the Markov source domain transitions fit observed target domain data ill-defined, as multiple real states with different dynamics can map to the same abstract state.
Consequently, naively attempting to correct $P_s(s_{i+1}^s|s_i^s, a_i)$ to approximate $\Pr(s_{i+1}^s|s_i^s, a_i)$ in the target domain will fail to ground the dynamics of the abstract state to the real world.

\begin{figure}[!htbp]
    \centering
    \includegraphics[width=\linewidth]{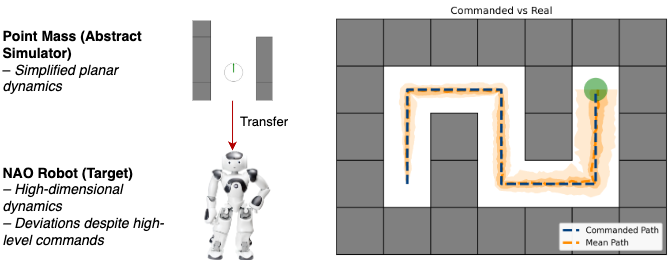}
    \caption{Identical velocity commands produce perfect tracking in point-mass simulation (blue) but significant deviations on the physical robot (orange: mean trajectory; shaded: 25-75\% and 5-95\% quantiles) due to unmodeled dynamics.}
    \label{fig:dynamics_mismatch}
    \vspace{-0.25cm}
\end{figure}


\section{ASTRA: Augmented Simulation with Self-Predictive Abstraction} 
\label{sec:astra}
A natural approach to abstract sim2real is to use real-world data to align the abstract simulator's dynamics with the real-world's dynamics. 
A straightforward way to address the partial observability induced by abstraction is to extend neural correction methods (e.g., \cite{hanna2021grounded,golemo2018sim}) with recurrent neural networks (RNNs) so that corrections can be based on full state-action histories.
Under this extension, these approaches would learn history-conditioned corrections that transform the simulator's predictions $s_{i+1}^s$ to better approximate the true abstract next state $\phi(s_{i+1}^t)$ observed in the real world.
This approach optimizes the hidden representation of the RNN solely for prediction accuracy through an MSE loss, which may not necessarily lead to a hidden representation that contains all task relevant information.
To explicitly shape the hidden state representation toward retaining task-relevant information, we introduce a new method, \textbf{ASTRA} (\textbf{A}ugmented \textbf{S}imulation with self‑predic\textbf{T}ive abst\textbf{RA}ction). 
The key novelty of ASTRA is to augment simulator grounding with loss terms motivated from the literature on self-predictive state abstractions.
Algorithm~\ref{alg:ais_training} presents the training procedure; Algorithm~\ref{alg:policy_learning} shows policy learning with the grounded simulator. Notably, ASTRA is agnostic to the choice of RL algorithm; we use PPO for all navigation tasks and NAO ball-kicking, SAC for humanoid locomotion and Ant low-level controller training.

\begin{figure*}[!t]
\centering
\includegraphics[width=0.95\textwidth]{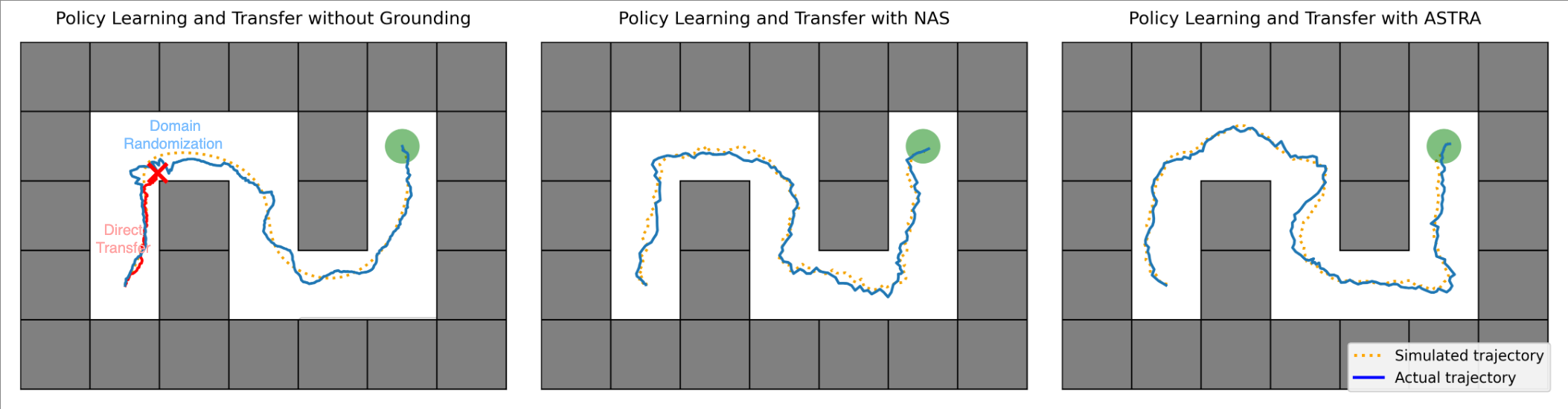}
\vspace{-10pt}
\caption{
Illustrations of trajectories learned by different approaches when transferring from PointMaze (abstract point-mass) to AntMaze (quadruped locomotion).
Orange dashed lines indicate trajectories generated in the abstract point-mass simulator; blue solid lines show deployment in the target AntMaze environment. 
\textbf{Left:} Without grounding, policies exploit the simplified dynamics to learn aggressive, near-optimal paths in simulation. Direct transfer (red) results in immediate collision at corners because the dynamics of quadrupedal locomotion differ from idealized point-mass movement. Domain randomization (blue) reaches the goal but fails to track the simulated trajectory.
\textbf{Center:} Incorporating history into grounding (using NAS \cite{golemo2018sim}) enables the PointMaze simulator to implicitly account for the dynamics of quadrupedal locomotion, resulting in policies that learn safer navigation patterns.  
\textbf{Right:} ASTRA learns policies that identify and navigate high-risk regions where abstract and real dynamics diverge most, resulting in higher success rate.
}
\label{fig:killer}
\vspace{-0.4cm}
\end{figure*}

\subsection{Grounding with Self-Prediction Losses}
ASTRA augments the abstract simulator with a learned latent dynamics model that preserves task-relevant information through self-predictive constraints. The approach consists of three neural components that share a common recurrent backbone:
\begin{itemize}
\item \textbf{Source history encoder} $\psi^s: \mathcal{H}^s \rightarrow \mathcal{Z}$: Maps abstract state-action histories to a latent space $\mathcal{Z}$.
\item \textbf{Latent dynamics model} $P_{\text{lat}}: \mathcal{Z} \times \mathcal{A} \rightarrow \Delta(\mathcal{Z})$: Predicts a distribution over next latent states given the current latent state and action.
\item \textbf{Reward predictor} $R_{\text{pred}}: \mathcal{Z} \times \mathcal{A} \rightarrow \mathbb{R}$: Estimates rewards from the target environment.
\item \textbf{Abstract state predictor} $f_{\text{abs}}: \mathcal{Z} \times \mathcal{A} \times \mathcal{S}^s \to \tilde{\mathcal{S}}^s$: Predicts next abstract states for simulator grounding.
\end{itemize}

Given an abstracted history $\Phi(h_i^t)$, the encoder produces a latent representation $z_i^s = \psi^s(\Phi(h_i^t)) \in \mathcal{Z}$. This encoding serves as the state representation for both simulator grounding and policy learning.
To train these components, we collect paired trajectories following the protocol from Golemo et al.~\cite{golemo2018sim}: 
for each transition $(s_i^t, a_i, s_{i+1}^t)$ collected from the target environment, we reset the abstract simulator to state $\phi(s_i^t)$ and execute the same action $a_i$ to obtain the simulator's prediction. 
For this paired data, we define the history abstraction operator:
$\Phi: \mathcal{H}^t \rightarrow \mathcal{H}^s$, where $\Phi(h^t_i) = (\phi(s_1^t), a_1, \phi(s_2^t), a_2, \ldots, \phi(s_i^t), a_{i-1})$
applies the abstraction function $\phi$ (from Section~\ref{sec:abstract-sim2real}) to each state in the target history. This operator constructs abstract histories that reflect real-world dynamics while maintaining the abstract state representation.
Our objective is to learn the encoder such that $z_i^s$ contains sufficient information both for grounding the abstract simulator and learning an effective policy.
Specifically, we seek a representation $z_i^s=\psi^s(\Phi(h_i^t))$ such that: 
(i) rewards are predictable from $z_i^s$, i.e., $\mathbb{E}[r_i^t \mid \Phi(h_i^t), a_i] \approx R_{\text{pred}}(z_i^s,a_i)$, and 
(ii) the next latent depends on the past only through $z_i^s$ (a Markov latent space), i.e., $P(z_{i+1}^s \mid \Phi(h_i^t), a_i) \approx P(z_{i+1}^s \mid z_i^s, a_i)$. 
We encourage $\psi^s$ to approximately satisfy these constraints via loss functions inspired by the concept of an \textit{approximate information state} (AIS) \cite{subramanian2022approximate,patil2024learning}. An AIS representation contains sufficient information to predict rewards in the real-world environment as well as the representation at the next time-step. AIS representations are closely related to self-predictive abstractions \cite{guo_byol-explore_2022,schwarzer_data-efficient_2021}. We will train the encoder used by ASTRA such that it produces an AIS hidden state representation.

Concretely, we optimize three complementary losses that correspond to the predictive components defined above.
First, the latent dynamics model, $P_{\text{lat}}$, outputs parameters $(\mu_i,\log\sigma_i^{2})$ of a Gaussian $\mathcal{N}(\mu_i,\sigma_i^{2})$ that approximates the distribution of $z_{i+1}^s$. We train both $P_{\text{lat}}$ and encoder $\psi^s$ to minimize the negative log-likelihood $\mathcal{L}_{\text{trans}} = -\sum_i\log\mathcal{N}(z_{i+1}^s \mid \mu_i,\sigma_i)$, where $z_{i+1}^s = \psi^s(\Phi(h_{i+1}^t))$. This drives $P_{\text{lat}}$ to mirror real-world transitions in latent space.
Second, the reward predictor $R_{\text{pred}}$ estimates target environment rewards. We jointly train $R_{\text{pred}}$ and $\psi^s$ to minimize $\mathcal{L}_{\text{rew}} = \sum_i\|R_{\text{pred}}(z_i^s,a_i)-r_i^t\|^{2}$.
Third, the abstract state predictor $f_{\text{abs}}$ uses the latent state representation as an input to correct the state transition of the abstract simulator so that it more accurately predicts the next abstract state in the real world
$
\mathcal{L}_{\text{abs}} = \sum_i\|f_{\text{abs}}(z_i^s,a_i,s^s_{i+1})-\phi(s^{t}_{i+1})\|^{2}.
$
ASTRA's total training objective combines these losses:
$
\mathcal{L} = \lambda_1\mathcal{L}_{\text{trans}} + \lambda_2\mathcal{L}_{\text{rew}} + \lambda_3\mathcal{L}_{\text{abs}}.
$
In our implementation, $\psi^s$, $P_{\text{lat}}$, $R_{\text{pred}}$, and $f_{\text{abs}}$ share a common GRU backbone with three task-specific output heads; we optimize all parameters jointly under $\mathcal{L}$.

\begin{algorithm}[!htb] 
\caption{ASTRA Simulator Grounding}
\label{alg:ais_training}
\begin{algorithmic}[1]
\FOR{epoch = 1 to $N_{\text{epochs}}$}
    \FOR{trajectory $(s^t_1, a_1, r^t_1, \ldots, s^t_T)$ in dataset}
        \FOR{$i = 1$ to $T-1$}
            \STATE $h^t_i = (s^t_1, a_1, \ldots, s^t_i, a_{i-1})$
            \STATE $z^s_i \gets \psi^s(\Phi(h^t_i))$
            \STATE $z^s_{i+1} \gets \psi^s(\Phi(h^t_{i+1}))$
            \STATE $(\mu_i,\log\sigma_i^2) \gets P_{\text{lat}}(z^s_i,a_i)$
            \STATE $\hat{r} \gets R_{\text{pred}}(z^s_i,a_i)$
            \STATE Get abstract simulator's next state $s^s_{i+1}$ from paired data
            \STATE $\hat{s}^s_{i+1} \gets f_{\text{abs}}(z^s_i,a_i,s^s_{i+1})$
            \STATE $\mathcal{L}_{\text{trans}} = -\log\mathcal{N}(z^s_{i+1} \mid \mu_i,\sigma_i^2)$
            \STATE $\mathcal{L}_{\text{rew}} = \|\hat{r}-r^t_i\|^{2}$
            \STATE $\mathcal{L}_{\text{abs}} = \|\hat{s}^s_{i+1}-\phi(s^t_{i+1})\|^{2}$
        \ENDFOR
        \STATE Update $(\psi^s, P_{\text{lat}}, R_{\text{pred}}, f_{\text{abs}})$ with gradient descent on $\mathcal{L}$
    \ENDFOR
\ENDFOR
\end{algorithmic}
\end{algorithm}

\subsection{Target Environment Abstraction}
ASTRA trains a policy that takes the learned AIS representation as input. Consequently, we need an encoder to produce these latent states when running the policy in the target environment.
To do this, ASTRA learns a target encoder $\psi^t:\mathcal{H}^t\rightarrow\mathcal{Z}$ that maps target histories into the same latent space as $\psi^s$, defining $z_i^t = \psi^t(h_i^t)$. To ensure compatibility, ASTRA enforces that $z_i^t$ and $z_i^s$ have similar distributions for corresponding actions. Let $p_i^s:= P(z_{i+1}^s|z^s_i, a_i)$ and $p_i^t := P(z_{i+1}^t|z^t_i, a_i)$. The alignment is achieved by minimizing:
$
\mathcal{L}_{\text{align}}=D(p_i^s, p_i^t)
$
where we use Maximum Mean Discrepancy (MMD) as $D$. 
For alignment, we freeze $\psi^s$, $P_{\text{lat}}$, $R_{\text{pred}}$ and update only $\psi^t$; after alignment, $\psi^t$ is kept fixed for deployment.
The target encoder then enables real-world deployment of  the policy trained in the grounded abstract simulator. 

\begin{algorithm}[!htb] 
\caption{Policy Learning with ASTRA-Grounded Simulator}
\label{alg:policy_learning}
\begin{algorithmic}[1]
\REQUIRE RL algorithm $\mathbb{A}$
\STATE Initialize $\mathbb{A}$
\FOR{episode = 1 to $M$}
    \STATE Initialize abstract simulator at $s_1^s$ and $h_1^s = (s_1^s)$
    \FOR{$i = 1$ to $T$}
        \STATE $z_i^s \gets \psi^s(h_i^s)$
        \STATE $a_i \sim \pi(z_i^s)$
        \STATE $z_{i+1}^s \sim P_{\text{lat}}(z_i^s, a_i)$
        \STATE $\hat{r}_i \gets R_{\text{pred}}(z_i^s, a_i)$
        \STATE Execute $a_i$ in abstract simulator
        \STATE $\hat{s}_{i+1}^s \gets f_{\text{abs}}(z_i^s, a_i, s^s_{i+1})$
        \STATE Set simulator state $s_{i+1}^s\gets\hat{s}_{i+1}^s$
        \STATE $h_{i+1}^s \gets (h_i^s, a_i, s_{i+1}^s)$
    \ENDFOR
    \STATE Update policy $\pi$ according to $\mathbb{A}$
\ENDFOR
\end{algorithmic}
\end{algorithm}

\section{Experiments}
\label{sec:experiments}
In this section, we empirically study abstract sim2real transfer to answer the following three questions: 
{(i) Does history-based grounding with recurrent policies enable transfer of policies trained in abstract simulators?}
(ii) How does the level of abstraction affect the relative importance of grounding methods versus domain randomization?
{(iii) Does learning a self-predictive state representation improve transfer efficacy compared to methods that only optimize abstract state prediction accuracy?}

To address these questions, we evaluate different sim2real methods in two real robot tasks and simulated navigation and humanoid locomotion with varying abstraction levels. 
We evaluate ASTRA and six baselines. Direct Transfer (DT) establishes the abstract sim2real gap. Domain Randomization (DR) is a basic application of randomization to the actions of the robot. COMPASS \cite{huang2023went} represents a strong domain-randomization style method. 
Rapid Motor Adaptation (RMA) \cite{kumar2021rma} uses history-based adaptation to infer environment physical parameters online, representing methods designed for parametric uncertainty (i.e. variations in known physical parameters).
In order to assess the necessity of learning a self-predictive state representation, we use neural-augmented simulation (NAS) \cite{golemo2018sim} as a representative neural grounding approach to sim2real.
NAS grounds the simulator via a recurrent correction function that is trained solely to predict the next abstract state that would occur in the real-world.
We also include IQL fine-tuning (DT+IQL) \cite{kostrikov2021offline}, which fine-tunes the DT policy on target-domain data to assess whether direct learning can match simulator grounding.
In our simulated experiments, we also consider directly training in the target environment.

\subsection{Legged Robot Navigation - Sim2Sim}
\label{sec:sim2sim_nav}
Our first sim2sim experiment uses two variants of AntMaze \cite{fu2020d4rl} as the target environment and a 2D point-mass abstract simulator. The abstract state space is the location and velocity of the agent whereas the full AntMaze state space includes a 29-dimensional state containing torso pose and joint angles/velocities.
To align the action-spaces, we use a separately trained low-level controller (frozen during experiments) to convert high-level velocity commands to joint torques in the target environment.
The abstraction gap is substantial as the simulator omits leg contacts, joint dynamics, and orientation drift that affect quadruped locomotion.

We evaluate on two maze configurations: U-Maze requiring a single $90^\circ$ turn and Long Maze with multiple turns. We evaluate methods with success rate over 100 evaluation trajectories and average performance over 10 seeds for each method. We sample initial and goal positions from Gaussians centered on fixed poses. For ASTRA and NAS, we collect 200 trajectories (average length 500 steps) using a random behavior policy $\pi_0$ to generate data for simulator grounding. For this domain, DR uses action noise ($\varepsilon \sim \mathcal{N}(0, 0.05)$) and scaling ($\delta \sim \mathcal{U}[-0.1, 0.1]$). COMPASS randomizes friction ($\mu \in [0.8, 1.2]$), position noise ($\pm 0.03$m), velocity noise ($\leq 0.02$m/s), heading bias ($\pm 5^\circ$), action parameters, and control timestep ($\Delta t \in [0.015, 0.025]$s). As upper bounds, we include policies trained directly in AntMaze: \emph{Target} uses the abstracted state $\phi(s^t)$ while \emph{Full Obs} has access to the complete state.

\begin{figure}[t]
  \centering
  \includegraphics[width=0.95\linewidth]{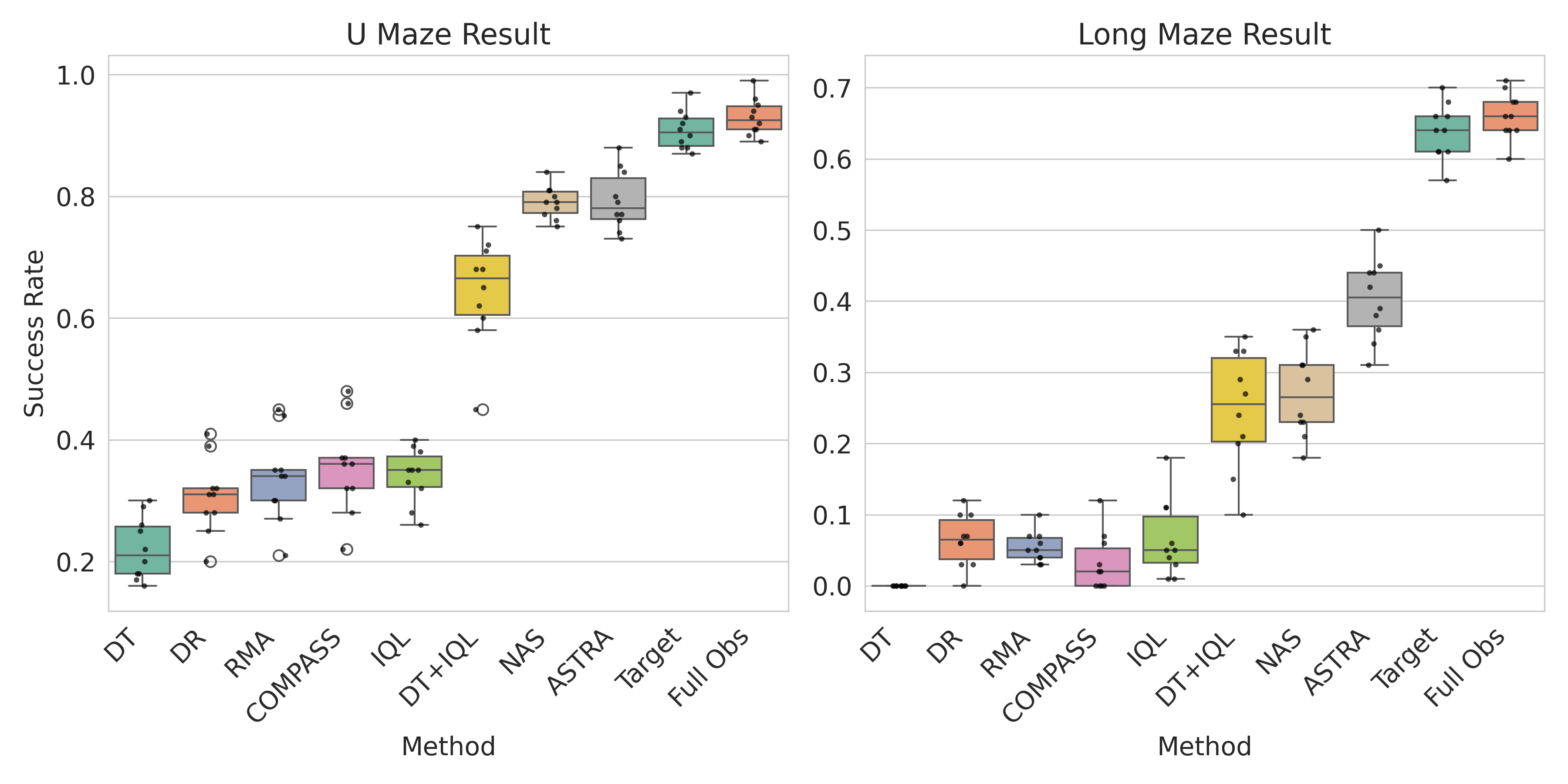}
  \caption{Success rates on U-Maze (left) and Long Maze (right) navigation tasks. ASTRA achieves the highest success rate in both settings. RMA performs between DR and COMPASS, indicating that adaptation methods designed for parametric uncertainty cannot fully address abstraction gaps. Direct IQL training on target data yields limited performance, while simulator pretraining followed by IQL fine-tuning (DT+IQL) approaches NAS but with higher variance.}
  \vspace{-5pt} 
  \label{fig:antmaze_results}
\end{figure}

The results shown in Figure~\ref{fig:antmaze_results} provide affirmative answers to our empirical questions.
Among all methods, ASTRA achieves the highest success rate in both maze configurations, demonstrating the effectiveness of self-predictive grounding for abstract sim2real.
RMA's performance between DR and COMPASS confirms that parametric adaptation cannot bridge abstraction gaps. While direct IQL training struggles on complex mazes, DT+IQL fine-tuning substantially improves. ASTRA’s consistent lead demonstrates that explicit dynamics grounding outperforms both parametric adaptation and direct fine-tuning. Furthermore, the learned grounding exhibits robustness to morphological variation: when evaluated on a modified Ant with leg segments scaled to $1.25\times$ their original length, the ASTRA policy trained on the standard Ant achieves 65\% success rate on U-Maze, compared to 21\% for direct transfer.

\subsection{Humanoid Locomotion}
\label{sec:humanoid}

\begin{figure}[htbp]
  \centering
  \includegraphics[width=0.40\textwidth]{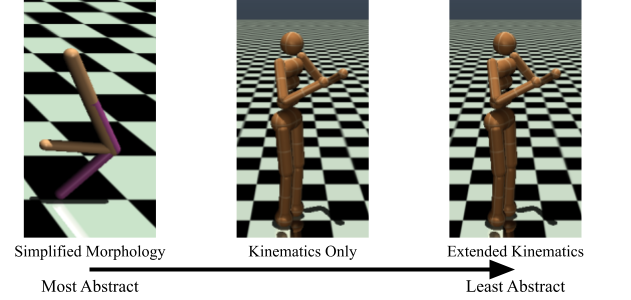}
  \caption{Abstraction hierarchy used for humanoid locomotion experiments: Walker2D, Kinematics, and Extended Kinematics.}
  \label{fig:humanoid_levels}
  \vspace{-5pt} 
\end{figure}

We next examine the influence of different levels of abstraction on abstract sim2real in simulated humanoid locomotion on the RL Humanoid benchmark \cite{duan2016benchmarking}.
We train policies that output joint position commands that a PD controller ($K_p=200$, $K_d=10$) converts to torques in the target environment.
We use the following abstract simulators with different levels of abstraction:
(1)~\textsc{Walker2D}:The most abstract variant models the humanoid's upper body as a single rigid link, reducing observations to leg positions, joint angles, and foot contacts while omitting arm, torso, and center-of-mass (CoM) information. The action space is 4-dimensional, controlling desired positions for four joints: left hip, left knee, right hip, and right knee, with each joint operating around a single axis. All other humanoid joints are fixed at constant values, effectively treated as rigid body components. A PD controller ($K_p=200$, $K_d=10$) converts the position commands to torques for these four active joints only.
(2)~\textsc{Kinematics}: Preserves the full humanoid morphology with observations including positions and velocities of all joints. 
The action space remains same as Walker2D, specifying desired positions for the same four joints (left hip, left knee, right hip, right knee), maintaining complete body structure while abstracting force-level dynamics.
(3)~\textsc{Extended Kinematics}: Augments kinematic abstraction with robot-level information including CoM and translational velocity~\cite{radosavovic2024real}, while maintaining the same 4-dimensional position control action space. This extension enables the policy to access whole-body state information that facilitates transfer to the full humanoid target environment.

We collect 200 trajectories (average length 500 steps) using a suboptimal PPO policy, as random policies fail immediately and thus provide low-relevance data for grounding. 
All methods in this experiment use GRU policies. As an upper-bound reference, we also train a policy directly in the target environment using the complete Humanoid observation space~\cite{duan2016benchmarking} while maintaining the same 4-dimensional action space, labeled as \emph{Target}. 
For humanoid domains, DR uses action noise ($\varepsilon \sim \mathcal{N}(0, 0.05)$) and scaling ($\delta \sim \mathcal{U}[-0.05, 0.05]$). COMPASS additionally randomizes joint friction ($\mu \in [0.8, 1.2]$), observation noise (scaling factor $\delta \sim \mathcal{U}[0.9, 1.1]$), and control timestep ($\Delta t \in [0.015, 0.025]$s).

\begin{figure}[t]
  \centering
  \includegraphics[width=\columnwidth]{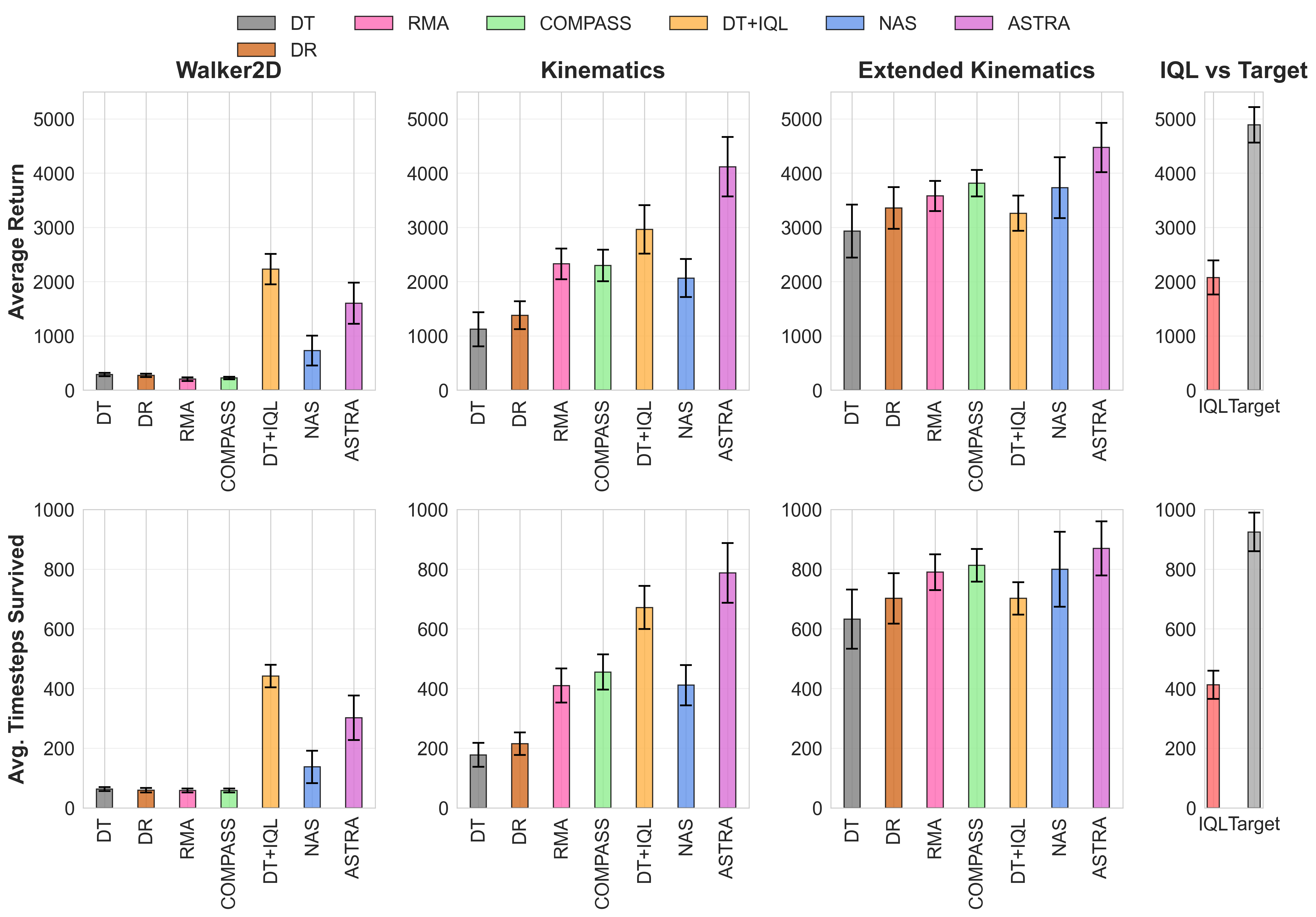}
  \caption{Humanoid locomotion results across three abstraction levels (10 seeds; higher is better). Left three columns compare all methods; rightmost column compares direct IQL training on target data versus training in the target environment (\emph{Target}).}
  \label{fig:humanoid_results}
  \vspace{-5pt} 
\end{figure}

Figure~\ref{fig:humanoid_results} reveals how abstraction level critically impacts transfer success. 
In the skeletal Walker2D setting, DT, DR, and COMPASS terminate quickly (within 63 timesteps). NAS shows improvement through history processing. ASTRA achieves the best performance, maintaining balance longest.
DT+IQL fine-tuning achieves strong performance in this extreme case, approaching direct IQL training on target data (rightmost column). However, as simulator fidelity increases, fine-tuning gains diminish: in \textsc{Kinematics} and \textsc{Extended Kinematics}, DT+IQL shows marginal improvement while ASTRA's advantage grows.
Full-body \textsc{Kinematics} stabilizes all algorithms and narrows the reality gap, with COMPASS now outperforming NAS. \textsc{Extended Kinematics} brings smaller gains, with NAS and COMPASS approaching the target-trained upper bound while ASTRA maintains its lead. These results demonstrate that (i) retaining essential information during abstraction is most effective for transfer, and (ii) {when abstraction is severe, ASTRA's self-predictive grounding proves most effective.

\subsection{Real Robot Evaluation}
We validate our approach on a physical NAO bipedal robot, testing transfer from highly abstract simulators to real hardware. The NAO presents unique challenges absent in simulation: imprecise odometry, foot slippage, actuator delays, and camera noise. We evaluate on two complementary tasks that stress different aspects of abstract sim2real transfer.

\subsubsection{NAO Navigation}
\label{sec:nao-nav}

The abstract simulator models the robot as a 2D point mass with velocity control. The real NAO executes commands through its walk engine, with observations from on-board localization including position, rotation, and velocity. The robot must navigate a physical maze to reach a 0.3-m radius goal zone without wall collisions. Runs are initialized from three distinct start poses per seed; episodes terminate after 500 control steps or upon completion. Performance metrics: success rate and distance traveled (m) over three seeds. Average distance is measured over successful trials, longer trajectories indicate that Agent learns more conservative, collision-avoiding behaviors.

For ASTRA and NAS, we augment 50 collected trajectories to 200 through rotational and translational transformations. 
For the DR baseline, we apply action noise ($\varepsilon \sim \mathcal{N}(0, 0.05)$) and scaling ($\delta \sim \mathcal{U}[-0.1, 0.1]$), consistent with the AntMaze configuration.
COMPASS randomizes ground friction $\mu \in \mathcal{U}[0.8, 1.2]$, foot slippage (20\% probability), position noise ($\pm 0.03$m), velocity noise ($\leq 0.02$m/s), heading bias ($\pm 5^\circ$), action noise ($\varepsilon \sim \mathcal{N}(0, 0.05)$, scaling $\delta \sim \mathcal{U}[-0.1, 0.1]$), and control timestep ($\Delta t \in \mathcal{U}[0.015, 0.025]$s).

As shown in the navigation results of Table~\ref{tab:nao_combined}, ASTRA achieves 73\% success rate, significantly outperforming all baselines (17\%--53\%). 
For successful trials, we also measure travel distance as a metric for both efficiency and risk, where shorter distances may indicate aggressive corner-cutting. We find that most baselines achieve lower distances than ASTRA but have lower success rates, suggesting they attempt to take corners too tightly and frequently fail. Because ASTRA implicitly accounts for the NAO’s low-level walking dynamics, it learns a more conservative policy that travels farther but succeeds more consistently.

\subsubsection{NAO Ball-Kicking}
\label{sec:nao-kick}

The NAO must kick a ball into a goal within 30s.
The abstract simulator models a 2D point agent and simplified ball physics. Real-world ball tracking comes from the robot's on-board perception and state-estimation modules which produce noisy position estimates. For each seed, we measure success rate over 20 trials with random starts.

We collect 200 trajectories using a random policy. COMPASS randomizes ground friction $\mu \in \mathcal{U}[0.8, 1.2]$, ball position/velocity (simulating state uncertainty), and post-contact ball direction (simulating ball-foot contact variations).

\begin{table}[htbp]
  \centering
  \caption{NAO robot evaluation (mean $\pm$ std, 3 seeds, 20 trials each)}
  \label{tab:nao_combined}
  \footnotesize
  \setlength{\tabcolsep}{4pt}
  \begin{tabular}{l|cc|c}
    \toprule
    & \multicolumn{2}{c|}{\textbf{Navigation Task}} & \textbf{Ball-Kicking Task} \\
    \textbf{Method} & Success Rate & Distance (m) & Success Rate \\
    \midrule
    DT       & 0.27 $\pm$ 0.21 & 10.91 & 0.07 $\pm$ 0.03 \\
    DR       & 0.33 $\pm$ 0.31 &  9.30 & 0.12 $\pm$ 0.04 \\
    RMA      & 0.51 $\pm$ 0.17 & 10.33 & 0.10 $\pm$ 0.01 \\
    DT+IQL   & 0.31 $\pm$ 0.10 & 10.72 & 0.08 $\pm$ 0.01 \\ 
    IQL      & 0.17 $\pm$ 0.14 &  9.77 & 0.05 $\pm$ 0.03 \\ 
    COMPASS  & 0.50 $\pm$ 0.08 & 12.70 & 0.40 $\pm$ 0.25 \\
    NAS      & 0.53 $\pm$ 0.06 & 12.41 & 0.37 $\pm$ 0.22 \\
    ASTRA    & \textbf{0.73 $\pm$ 0.05} & 12.33 & \textbf{0.56 $\pm$ 0.05} \\
    \bottomrule
  
  \end{tabular}
  \vspace{-5pt}
\end{table}

For the ball-kicking task of table~\ref{tab:nao_combined}, right, ASTRA achieves 56\% success rate, substantially outperforming all baselines. NAS (37\%) shows improvement over DT and DR through history processing, but still falls short of ASTRA's performance in handling camera noise and contact uncertainty.

\subsection{Component Analysis}
To isolate each training objective's contribution, we ablate ASTRA's loss components on the \textsc{Long Maze} navigation task: latent dynamics prediction ($\mathcal{L}_\text{trans}$), reward prediction ($\mathcal{L}_\text{rew}$), and abstract state correction ($\mathcal{L}_\text{abs}$). Since $\mathcal{L}_\text{abs}$ provides the essential grounding signal, we evaluate Full ASTRA, ASTRA without $\mathcal{L}_\text{trans}$, and ASTRA without $\mathcal{L}_\text{rew}$. Using $\mathcal{L}_\text{abs}$ alone is equivalent to NAS.

\begin{table}[htbp]
  \centering
  \caption{Component ablation (mean $\pm$ std over 10 seeds)}
  \label{tab:ablation}
  \footnotesize
  \setlength{\tabcolsep}{5pt}
  \begin{tabular}{l|c}
    \toprule
    \textbf{Configuration} & \textbf{Success Rate} \\
    \midrule
    Full ASTRA                                      & \textbf{0.40 $\pm$ 0.05} \\
    \quad w/o dynamics ($\mathcal{L}_\text{trans}$) & 0.35 $\pm$ 0.04 \\
    \quad w/o reward ($\mathcal{L}_\text{rew}$)     & 0.29 $\pm$ 0.10 \\
    NAS                                             & 0.27 $\pm$ 0.06 \\
    \bottomrule
  \end{tabular}
  \vspace{-5pt} 
\end{table}

Table~\ref{tab:ablation} shows that both auxiliary objectives contribute to ASTRA's performance, with reward prediction playing the dominant role. Removing $\mathcal{L}_\text{rew}$ causes a substantial drop (from 0.40 to 0.29) with doubled variance, indicating that task-relevant reward signals are critical for learning stable, control-oriented representations. Removing $\mathcal{L}_\text{trans}$ yields a smaller decrease (to 0.35) with stable variance, suggesting that self-predictive dynamics provide complementary regularization. The gap between Full ASTRA and NAS (0.40 vs 0.27) validates that both objectives contribute substantially beyond pure state prediction.

\subsection{Data Efficiency Analysis}
Finally, we examine how performance scales with dataset size using PointMaze-to-AntMaze transfer as a representative task.
We evaluate ASTRA's data efficiency compared to the strongest baseline (NAS) across six dataset sizes:
25\%, 50\%, 75\%, 100\%, 125\%, and 150\% of a baseline dataset containing 200 trajectories. Each configuration is evaluated through downstream RL success rate across 5 independent seeds on the Long-Maze task.
Figure~\ref{fig:data_efficiency} shows the position coverage (top), velocity coverage (middle), and transfer performance (bottom) as dataset size increases. The results demonstrate clear diminishing returns in data collection for simulator grounding. 
ASTRA achieves its steepest improvement between 25\% and 75\% of the baseline dataset, with performance plateauing beyond 100\%. This trend also holds for NAS, though at lower absolute performance.
Notably, doubling the dataset from 75\% (150 trajectories) to 150\% (300 trajectories) yields less than 10\% improvement in success rate. The variance patterns indicate that performance stability emerges around 100\% data, suggesting this represents sufficient coverage of the state space.

\begin{figure}[t]
  \centering
  \includegraphics[width=\columnwidth]{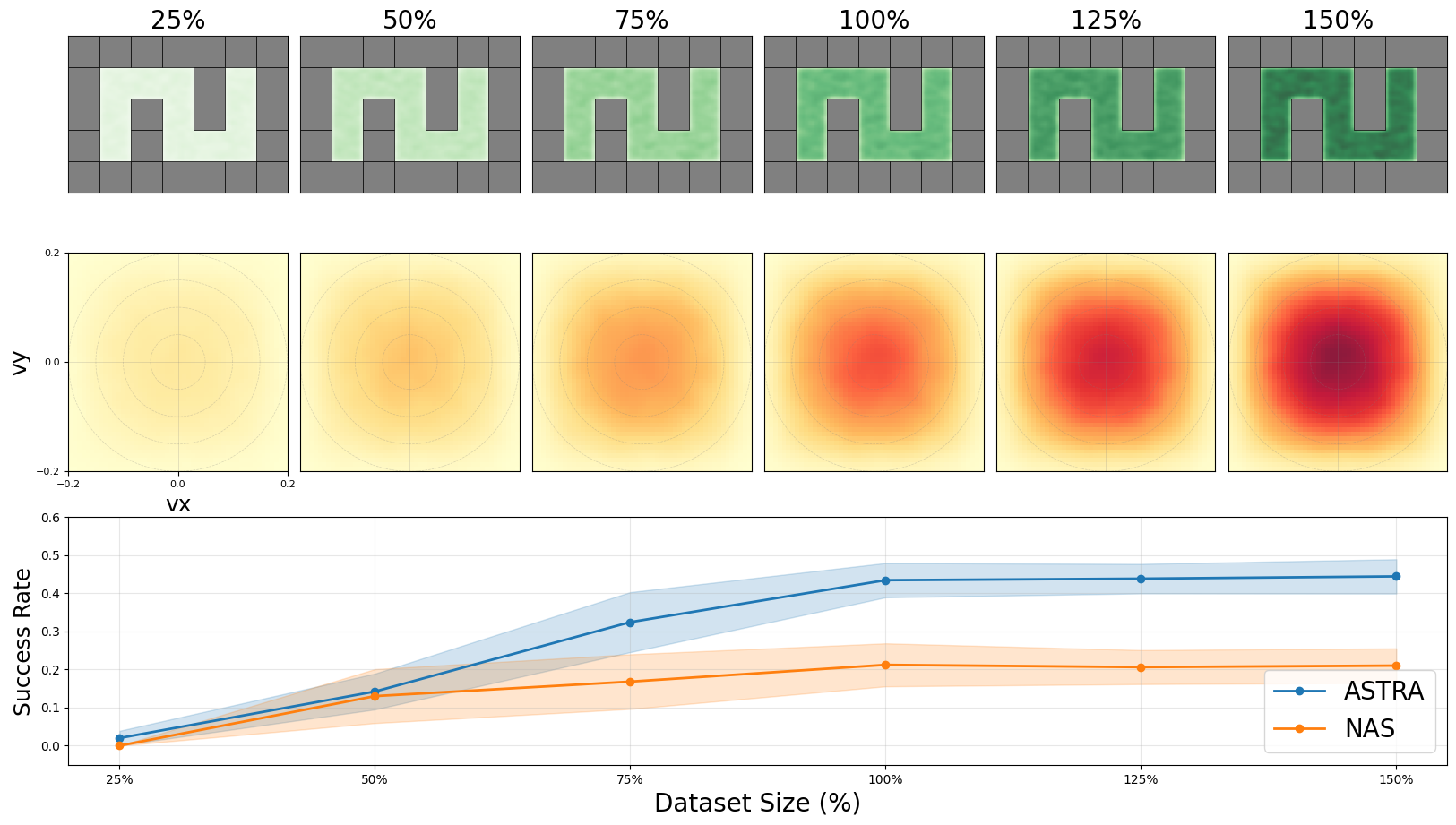}
    \caption{Dataset efficiency analysis. Top: position coverage heatmap showing visited states. Middle: velocity distribution histogram. Bottom: transfer performance showing mean $\pm$ standard deviation across 5 seeds (solid line: mean, shaded region: $\pm 1$ std).}
  \label{fig:data_efficiency}
  \vspace{-5pt} 
\end{figure}

\section{Conclusion}
In this paper, we formalized the abstract sim2real problem, which highlighted the need to learn history-based policies and grounding corrections that account for partial observability induced by abstraction. Based on this formalism, we introduced ASTRA, a method that learns a correction function for the abstract simulator to implicitly capture the impact of variables that were abstracted away. Both sim2sim and sim2real experiments demonstrate that ASTRA enables successful policy transfer from abstract simulators to target domains, with learned grounding showing robustness even under morphological variation of the target robot.

We note a few limitations as a basis for future work. History-based grounding may be insufficient if too much abstraction is applied; identifying acceptable abstraction levels is an interesting future direction. ASTRA requires a known state mapping, limiting applicability to higher-level variables rather than raw sensors, and grounding still requires non-trivial real-world data, though less than direct policy learning. Since all simulators are abstract to some degree \cite{hofer_perspectives_2020}, future work could provide a more unifying analysis beyond the more abstract end of the spectrum studied here.

\bibliographystyle{IEEEtran}
\bibliography{ref}

\end{document}